\documentclass{article}

\usepackage{microtype}
\usepackage{graphicx}
\usepackage{subcaption}
\usepackage{booktabs} %

\usepackage{hyperref}

\usepackage[preprint]{icml2026}

\usepackage{amsmath}
\usepackage{amssymb}
\usepackage{mathtools}
\usepackage{amsthm}

\usepackage{multirow}
\usepackage[table,svgnames]{xcolor}
\usepackage[zerostyle=b,scaled=0.95]{newtxtt}
\usepackage{xspace}
\usepackage{xurl}

\definecolor{lightred}{RGB}{245, 216, 216}

\usepackage[breakable]{tcolorbox}

\usepackage{mdframed}
\usepackage{tcolorbox}
\usepackage{listings}
\usepackage{makecell}
\usepackage{fontawesome5}
\usepackage{marvosym}

\usepackage[capitalize,noabbrev]{cleveref}

\theoremstyle{plain}

\theoremstyle{definition}

\theoremstyle{remark}

\newcommand{\hflogo}{\raisebox{-0.25\height}{\includegraphics[height=1.45em]{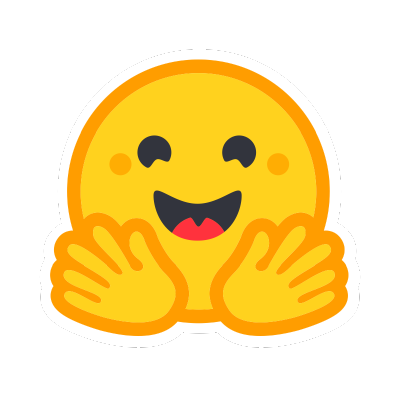}}}
\newcommand{\ghlogo}{\raisebox{-0.2\height}{\includegraphics[height=1.3em]{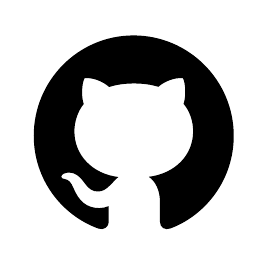}}}

\newcommand{\ours}{\textsc{WideSeek-R1}\xspace}

\icmltitlerunning{\ours: Exploring Width Scaling for Broad Information Seeking via MARL}

\begin{document}

\twocolumn[
  \icmltitle{\ours: Exploring Width Scaling for Broad \\ Information Seeking via Multi-Agent Reinforcement Learning}

  \icmlsetsymbol{equal}{*}
  \icmlsetsymbol{corr}{$\dagger$}

  \vspace{-1mm}
  \begin{icmlauthorlist}
    \icmlauthor{Zelai Xu}{thu,equal}
    \icmlauthor{Zhexuan Xu}{thu,equal}
    \icmlauthor{Ruize Zhang}{sigs,equal}
    \icmlauthor{Chunyang Zhu}{inf}
    \icmlauthor{Shi Yu}{iiis}
    \\
    \icmlauthor{Weilin Liu}{inf}
    \icmlauthor{Quanlu Zhang}{inf}
    \icmlauthor{Wenbo Ding}{sigs}
    \icmlauthor{Chao Yu}{sigs,corr}
    \icmlauthor{Yu Wang}{thu,corr}
    \\
    \vspace{0.5em}
    \href{https://wideseek-r1.github.io/}{{\color{black}\faGlobe}~\,Project Page} \quad
    \href{https://rlinf.readthedocs.io/en/latest/rst_source/examples/agentic/wideseek_r1/index.html}{{\color{black}\faBookOpen}~\,Docs} \quad
    \href{https://github.com/RLinf/RLinf/tree/main/examples/agent/wideseek_r1}{{\color{black}\ghlogo}~Code} \quad
    \href{https://huggingface.co/datasets/RLinf/WideSeek-R1-train-data}{{\color{black}\faDatabase}~\,Dataset} \quad
    \href{https://huggingface.co/RLinf/WideSeek-R1-4b}{{\color{black}\hflogo}~Model}
  \end{icmlauthorlist}

  \icmlaffiliation{thu}{EE, Tsinghua University}
  \icmlaffiliation{sigs}{SIGS, Tsinghua University}
  \icmlaffiliation{iiis}{IIIS, Tsinghua University}
  \icmlaffiliation{inf}{Infinigence AI}

  \icmlcorrespondingauthor{Zelai Xu}{zelai.eecs@gmail.com}
  \icmlcorrespondingauthor{Chao Yu}{yuchao@sz.tsinghua.edu.cn}
  \icmlcorrespondingauthor{Yu Wang}{yu-wang@tsinghua.edu.cn}


  \vskip 0.3in
]

\printAffiliationsAndNotice{\icmlEqualContribution\icmlCorrespondingAuthor}

\begin{abstract}
Recent advancements in Large Language Models (LLMs) have largely focused on depth scaling, where a single agent solves long-horizon problems with multi-turn reasoning and tool use. However, as tasks grow broader, the key bottleneck shifts from individual competence to organizational capability. In this work, we explore a complementary dimension of width scaling with multi-agent systems to address broad information seeking. Existing multi-agent systems often rely on hand-crafted workflows and turn-taking interactions that fail to parallelize work effectively. To bridge this gap, we propose \ours, a lead-agent--subagent framework trained via multi-agent reinforcement learning (MARL) to synergize scalable orchestration and parallel execution. By utilizing a shared LLM with isolated contexts and specialized tools, \ours jointly optimizes the lead agent and parallel subagents on a curated dataset of 20k broad information-seeking tasks. Extensive experiments show that \ours-4B achieves an item F1 score of 40.0\% on the WideSearch benchmark, which is comparable to the performance of single-agent DeepSeek-R1-671B. Furthermore, \ours-4B exhibits consistent performance gains as the number of parallel subagents increases, highlighting the effectiveness of width scaling.

\end{abstract}

\section{Introduction}
\label{sec:intro}
\begin{figure}[t]
\centering
\includegraphics[width=\linewidth]{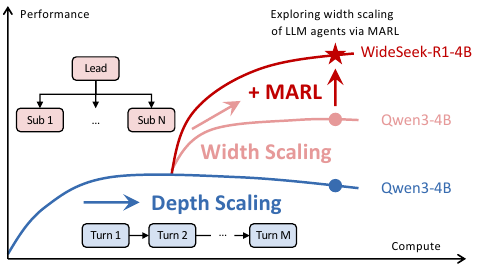}
\caption{Comparison of depth and width scaling. While depth scaling enhances performance through sequential multi-turn interactions, width scaling orchestrates multi-agent systems for parallel execution. \ours pushes the frontier of width scaling via MARL for synergized orchestration and execution.}
\label{fig:teaser}
\end{figure}

Recent advances in Large Language Models (LLMs)~\cite{guo2025deepseek,team2025kimi,google2025gemini3} have significantly improved single-agent capabilities in multi-turn reasoning and tool use.
Existing efforts mainly focus on \textit{depth scaling}, characterized by extended chain-of-thought and sequential actions to address long-horizon problems.
However, as tasks expand in breadth, the primary bottleneck shifts from individual competence to organizational capability~\cite{kimiteam2026kimi}.
This motivates a complementary dimension of \textit{width scaling} with multi-agent systems, where a lead agent decomposes broad objectives into independent subtasks and orchestrates parallel subagents to tackle problems beyond the capacity of a single agent.

Broad information seeking~\cite{wong2025widesearch} serves as an ideal testbed to explore the width scaling dimension.
Unlike deep research~\cite{mialon2023gaia,wei2025browsecomp} that requires an intensive investigation of a single complex query, broad information seeking involves a wide range of subtasks to gather and synthesize attributes of multiple entities into a structured tabular format.
Single-agent methods suffer from two limitations in such scenarios.
First, context pollution~\cite{anthropic2026building} degrades the agent's performance as its context accumulates irrelevant information from previous subtasks.
Second, sequential execution restricts efficiency by forcing the agent to process independent subtasks serially.
These limitations underscore the necessity of multi-agent systems, which naturally enable context isolation and parallel execution for effective width scaling.

However, existing multi-agent systems have yet to fully realize the potential of width scaling, primarily because few systems are trained end-to-end to learn scalable orchestration and parallel execution.
At the orchestration level, most prior work~\cite{li2023camel,wu2024autogen} relies on hand-crafted workflows rather than learned agents, hindering flexible and scalable coordination of multiple agents.
At the execution level, current systems~\cite{hu2025owl,li2025flow} typically process subtasks one-at-a-time and adopt turn-taking interactions that serialize progress and fail to parallelize subtasks.
As a result, the performance of multi-agent systems is bottlenecked by limited scalability and insufficient parallelization.

To bridge this gap, we introduce \ours, a lead-agent--subagent system trained via multi-agent reinforcement learning (MARL) to synergize scalable orchestration and parallel execution for broad information seeking.
We instantiate the lead agent and the subagents using a shared LLM with different tools and isolated contexts.
The lead agent focuses on task decomposition and multi-turn orchestration with a single tool named \texttt{call\_subagent} to delegate subtasks.
Each subagent then executes the assigned subtask in parallel by utilizing \texttt{search} and \texttt{access} tools to gather information and return its findings.
To enable multi-agent learning beyond multi-hop QA datasets, we construct a training set of 20k broad information-seeking tasks.
Using trajectories from both the lead agent and the subagents, we train a \ours-4B model via MARL to jointly optimize scalable orchestration and parallel information seeking.

Extensive experiments are conducted to demonstrate that \ours pushes the boundaries of width scaling. 
On the WideSearch benchmark, \ours-4B achieves an item F1 score of 40.0\%, which is comparable to single-agent DeepSeek-R1-671B and significantly outperforms multi-agent 8B baselines.
Furthermore, we investigate the scaling properties of both depth and width dimensions.
While depth scaling quickly reaches a plateau, \ours-4B exhibits continuous performance gains as the number of parallel subagents increases.
We also evaluate our method on standard QA benchmarks and perform ablation studies on learning agents and training data to validate that MARL synergizes orchestration and parallel information seeking.

In summary, our contributions are threefold:
\begin{itemize}
    \item We introduce \ours, a multi-agent system trained via MARL to synergize scalable orchestration and parallel execution for broad information seeking.
    \item We open-source a large-scale dataset of 20k broad information-seeking tasks, offering a complementary training resource to existing multi-hop datasets.
    \item We demonstrate the effectiveness of width scaling with \ours-4B, which achieves comparable performance to the DeepSeek-R1-671B and exhibits consistent gains as the number of parallel agents increases.
\end{itemize}

\section{Related Work}
\label{sec:related}
\textbf{Scaling Dimensions in LLMs.}
The evolution of LLMs has been primarily driven by two scaling paradigms: training-time scaling and test-time scaling.
Training-time scaling enhances foundational capabilities of LLMs by increasing model parameters, dataset size, and total training compute~\cite{kaplan2020scaling,hoffmann2022training}, whereas test-time scaling boosts performance by allocating more compute at inference time.
A prominent line of work scales depth at test time, where reasoning models~\cite{jaech2024openai,guo2025deepseek} and agentic models~\cite{team2025kimi,google2025gemini3} leverage extended chain-of-thought and multi-turn tool use to solve long-horizon problems.
Our work investigates a complementary dimension of width scaling with multi-agent systems.
While prior work attains width-like gains through sampling-based aggregation, such as best-of-$N$~\cite{gao2023scaling} and self-consistency~\cite{wang2022self}, these methods typically improve reliability by repeatedly sampling solutions for the same task.
In contrast, our work decomposes a broad objective into independent subtasks and improves performance by scalable orchestration and parallel execution.

\textbf{Search Agents and Systems.}
Building autonomous agents for information seeking has transitioned from simple retrieval-augmented generation to complex, multi-step reasoning. 
Single-agent methods, such as Search-R1~\cite{jin2025search} and ASearcher~\cite{gao2025beyond}, leverage reinforcement learning (RL) to optimize multi-turn tool use in open-ended environments.
While effective for deep, multi-hop queries, these methods suffer in broad information-seeking tasks due to context pollution and sequential execution.
To address these bottlenecks, multi-agent frameworks like CAMEL~\cite{li2023camel} and AutoGen~\cite{wu2024autogen} have been proposed to decompose complex tasks into structured workflows.
However, existing work typically relies on hand-crafted workflows, which limit the flexibility and scalability of the systems.
Our work differs from these frameworks by employing end-to-end training via MARL to incentivize scalable orchestration of parallel agents.

\begin{figure*}[h]
\centering
\includegraphics[width=\linewidth]{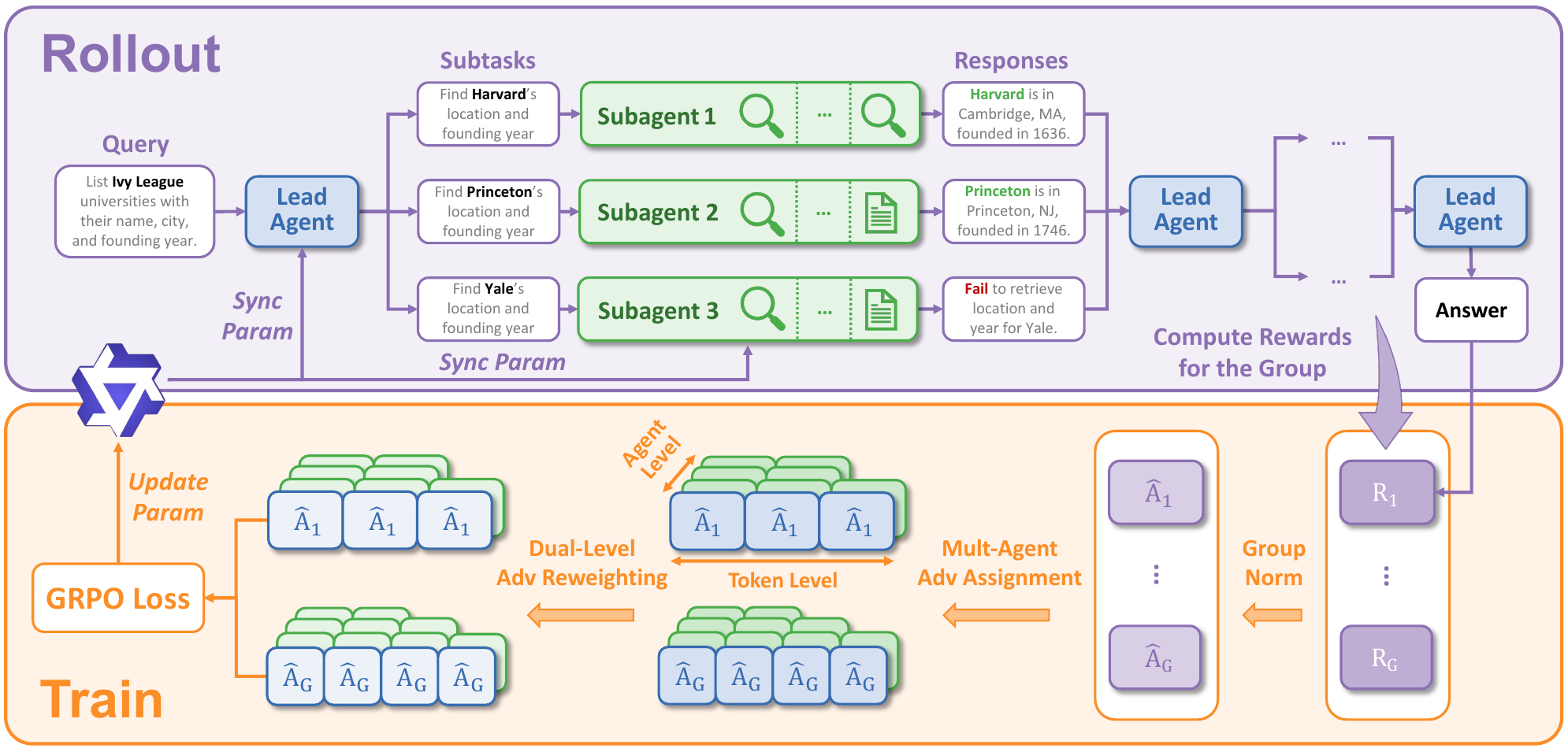}
\caption{Overview of \ours Rollout and Training Pipeline. (1) \textbf{Rollout}: The lead agent coordinates task decomposition while subagents execute parallel subtasks using external tools. (2) \textbf{Training}: We adopt group-level advantage normalization and assign the same advantage to all agents within each multi-agent system, followed by a dual-level advantage reweighting mechanism at both token level and agent level applied to the GRPO objective for effective multi-agent, multi-turn RL training.}
\label{fig:overview}
\end{figure*}

\textbf{Agentic RL for LLMs.}
The remarkable success of reasoning RL for LLMs~\cite{guo2025deepseek} has catalyzed the development of agentic RL, where models are trained to master tool-use and solve long-horizon problems, such as search~\cite{jin2025search}, code generation~\cite{wei2025swe}, and computer use~\cite{wang2025ui}.
However, multi-agent RL for LLMs remains relatively underexplored.
One line of work~\cite{zhao2025absolute,yuan2025marshal} focuses on training fully distributed systems with self-play RL to incentivize reasoning capability.
Another line of work~\cite{hu2025owl,li2025flow} considers hierarchical systems and trains agents with different roles.
Our work differs from these multi-agent systems in two aspects.
First, unlike prior work that trains part of the agents or uses separate models, we jointly optimize the lead agent and the subagents with a shared model.
Second, existing systems typically adopt turn-taking interaction that processes subtasks one-at-a-time, while our method enables parallel execution of subtasks to explore the potential of width scaling.

\section{\ours}
\label{sec:method}

In this work, we introduce \ours, a hierarchical lead-agent--subagent system trained via MARL to synergize scalable orchestration and parallel execution for width scaling.
As shown in Fig.~\ref{fig:overview}, we instantiate the lead agent and the subagents using a shared LLM with isolated contexts and specialized tools.
The lead agent focuses on task decomposition and multi-turn orchestration, while each subagent executes assigned subtasks in parallel by utilizing external tools to gather information and return findings.
We jointly optimize the lead agent and subagents via end-to-end MARL, enabling the system to learn effective coordination and information seeking simultaneously.

\subsection{Lead Agent for Scalable Orchestration}
\label{sec:method:lead_agent}

The lead agent is responsible for decomposing a broad task into parallelizable subtasks and delegating them to subagents.
Unlike existing multi-agent systems that rely on hand-crafted workflows, our lead agent is trained to perform scalable and learnable orchestration, enabling flexible coordination as the number of subagents increases.

The only tool available to the lead agent is \texttt{call\_subagent}, which we intentionally restrict to avoid context pollution.
In each turn, the lead agent invokes this tool to generate a set of well-defined subtasks, each accompanied by a clear prompt that serves as task guidance, and assigns them to subagents for parallel execution.
The lead agent remains idle until all subagents complete their respective subtasks, after which it proceeds to the next turn.
This process continues until the final turn, which produces the complete answer.
An effective lead agent must not only decompose a broad task into manageable subtasks that can be solved in parallel, but also formulate clear and informative prompts for subagents, as these prompts serve as their primary source of instruction.

\subsection{Subagents for Parallel Execution}
\label{sec:method:subagent}

The subagents are responsible for parallel information seeking, enabling width scaling by executing multiple subtasks simultaneously.
This design addresses the context pollution and sequential execution bottlenecks that plague single-agent methods in broad information-seeking tasks.

Upon assignment by the lead agent, each subagent operates in parallel within an isolated context, employing multi-turn tool-integrated reasoning to execute its specific subtask. 
The subagents are equipped with two tools: (1) \texttt{search}, which retrieves relevant snippets and URLs for a given query; and (2) \texttt{access}, which generates a summary from a specific URL conditioned on the given query.
Once all parallel threads conclude, control reverts to the lead agent for next decomposition.
In this framework, subagents function as high-level tools for the lead agent, where their precision in filtering and synthesizing external information is paramount to the system's overall performance.

\subsection{Multi-Agent Reinforcement Learning}
\label{sec:method:MARL}

We jointly optimize the lead agent and subagents through end-to-end multi-agent reinforcement learning (MARL) with a shared model, enabling the simultaneous learning of orchestration and information-seeking behaviors.
Our method builds upon GRPO~\cite{shao2024deepseekmath} and extends it for multi-agent systems with two key designs: multi-agent advantage assignment and dual-level advantage reweighting.

\textbf{Training Objective.}
For each query $q\!\sim\!\mathcal{D}$, a group of $G$ multi-agent rollouts $\{\tau_i\}_{i=1}^{G}$ is sampled with policy $\pi_{\theta_{\mathrm{old}}}$. Rollout $\tau_i$ contains $N_i$ agents, indexed by $a \in \{1,\dots,N_i\}$. Agent $a$ in rollout $i$ produces a multi-turn trajectory with $T_{i,a}$ turns. At turn $t$, the agent outputs a token sequence $o_{i,a}^{t}$ of length $\lvert o_{i,a}^{t} \rvert$, where the $j$-th token in the sequence is denoted by $o_{i,a}^{t,j}$.
 Our training objective is
\begin{equation}
\label{eq:loss}
\mathbb{E}
\Bigg[
\frac{1}{G}\!
\sum_{i=1}^{G}
{\color{DarkRed}\frac{1}{N_i}\!
\sum_{a=1}^{N_i}}\
{\color{DarkBlue}\frac{1}{\sum_{t=1}^{T_{i,a}} \lvert o^t_{i,a} \rvert}\!
\sum_{t=1}^{T_{i,a}}\!
\sum_{j=1}^{\lvert o^t_{i,a} \rvert}}
\mathcal{L}_\theta\!\left( r^{t,j}_{i,a},\,\hat{A}_i\right)\!
\Bigg],
\end{equation}

where the expectation is over $q \sim \mathcal{D}$ and $\{ \tau_i \}_{i=1}^{G} \sim \pi_{\theta_{\mathrm{old}}}(\cdot \mid q)$, the clipped policy gradient loss $\mathcal{L}_\theta\big( r,\hat{A}\big)$ is

\begin{equation}
\min\!\Big(
r(\theta)\,\!\hat{A},\;
\!\operatorname{clip}\!\big(r(\theta),\,\!1\!-\!\epsilon_{\text{low}},\,\!1\!+\!\epsilon_{\text{high}}\big)\,\!\hat{A}
\Big),
\end{equation}

and the importance ratio $r^{t,j}_{i,a}(\theta)$ is

\begin{equation}
\frac{\pi_\theta\!\left(o^{t,j}_{i,a} \mid s^t_{i,a},\, o^{t,<j}_{i,a}\right)}
{\pi_{\theta_{\mathrm{old}}}\!\left(o^{t,j}_{i,a} \mid s^t_{i,a},\, o^{t,<j}_{i,a}\right)}.
\end{equation}

\textbf{Multi-Agent Advantage Assignment.}
Credit assignment across multiple agents is a challenge unique to multi-agent settings, as agents can affect the final outcome both directly and indirectly.
To ensure training stability and prevent reward hacking, we use a verifiable outcome reward $R_i$ for each multi-agent rollout $\tau_i$, where $R_i$ is primarily determined by the answer's consistency with the ground truth.
We then compute a group-normalized advantage $\hat{A}_i = (R_i - \mu) / \sigma$ across the $G$ rollouts in the same group, where $\mu$ and $\sigma$ are the mean and standard deviation of rewards within the group.
This simple yet effective approach extends GRPO to multi-agent systems: the same advantage $\hat{A}_i$ is assigned to all agents and all tokens in the same multi-agent rollout, enabling joint optimization without complex credit assignment that may lead to reward hacking.

\textbf{Dual-Level Advantage Reweighting.}
We introduce a dual-level advantage reweighting mechanism within the policy gradient objective to better handle multi-agent, multi-turn training of LLMs.

\begin{itemize}
    \item \textit{Token-level reweighting across turns.} Following DAPO~\cite{yu2025dapo}, we reweight advantages by averaging over all tokens produced by an agent across all turns, as highlighted in blue in Eq.~(\ref{eq:loss}).
    This ensures that turns with more tokens have greater influence on the training loss in multi-turn settings, rather than being diluted by turn-level averaging in standard GRPO.    

    \item \textit{Agent-level reweighting.} We further reweight advantages by averaging over the agents in each multi-agent rollout, as highlighted in red in Eq.~(\ref{eq:loss}).
    This prevents rollouts with many subagents from dominating the gradient and reduces a common failure mode where the lead agent repeatedly spawn subagents without improving answer quality. With agent-level averaging, adding agents only helps if it improves the final reward.
\end{itemize}

Further details regarding the notation, state representation, tool-call handling, and the full reward formulation can be found in Appendix~\ref{appendix:rollout}.

\section{Training Data Construction}
\label{sec:data}
\begin{figure*}[t]
\centering
\includegraphics[width=\linewidth]{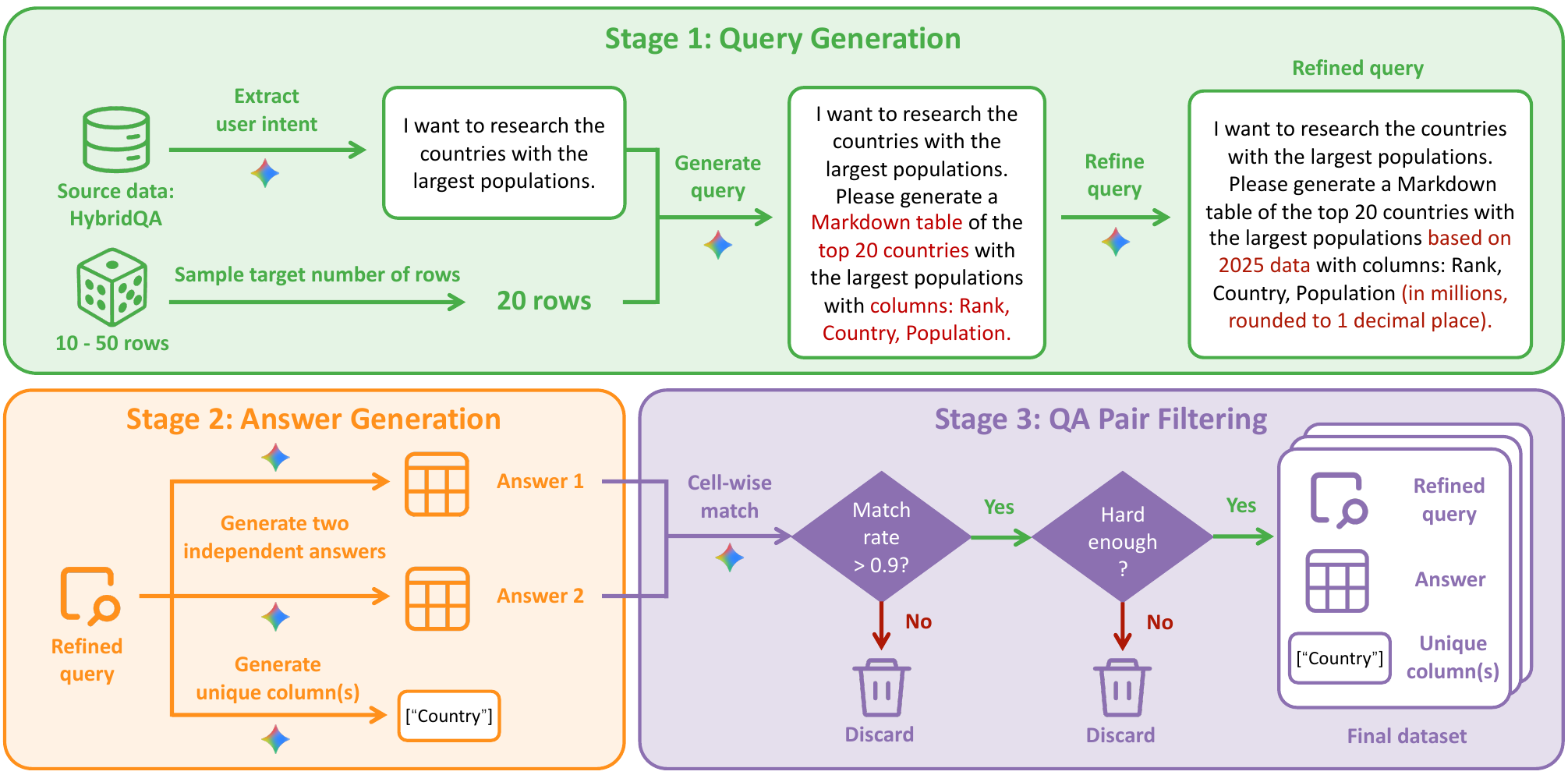}
\caption{Overview of our Automated Data Construction Pipeline. The pipeline comprises three stages: 
(1) \textbf{Query Generation}, where we extract user intents from HybridQA~\cite{chen2020hybridqa} and refine them into complex, schema-constrained queries that mandate specific table structures and broad coverage;
(2) \textbf{Answer Generation}, where we prompt the model to generate two responses independently along with the unique column(s), enabling self-consistency verification; and
(3) \textbf{QA Pair Filtering}, where we rigorously screen the data by discarding instances with low consistency or insufficient difficulty, ensuring that only robust and challenging samples remain in the final dataset.
\raisebox{-0.6ex}{\includegraphics[height=2.4ex]{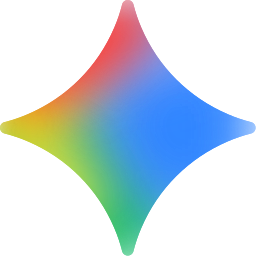}} marks the steps powered by the \texttt{gemini-3-pro-preview} API.}

\label{fig:data_pipeline}
\end{figure*}

\begin{table*}[t]
    \centering
    \caption{
    Results on WideSearch. We report Avg@4 and Max@4 for Item and Row F1 scores, and Avg@4 and Pass@4 for Success Rate. 
    \ours-4B outperforms all 4B and 8B baselines on five out of six metrics. 
    Notably, \ours-4B achieves performance comparable to the single-agent DeepSeek-R1-671B while utilizing nearly 170$\times$ fewer parameters.
    }
    \setlength{\tabcolsep}{8pt}
\begin{tabular}{cccccccc}
\toprule
\multirow{2}{*}{\textbf{Setting}} & \multirow{2}{*}{\textbf{Model}} & \multicolumn{2}{c}{\textbf{Item F1 Score (\%)}} & \multicolumn{2}{c}{\textbf{Row F1 Score (\%)}} & \multicolumn{2}{c}{\textbf{Success Rate (\%)}} \\
\cmidrule(lr){3-4} \cmidrule(lr){5-6} \cmidrule(lr){7-8}
 & & Avg@4 & Max@4 & Avg@4 & Max@4 & Avg@4 & Pass@4 \\
\midrule

\multirow{5}{*}{\textit{\shortstack{Single\\Agent}}} 
 & SingleSeek-R1-4B
 & 28.1 & 39.2 & 6.5 & 12.5 & 0.3 & 1.0 \\
 & Qwen3-4B
 & 20.1 & 30.2 & 3.0 & 4.8 & 0.0 & 0.0 \\
 & Search-R1-7B
 & 15.5 & 24.4 & 2.0 & 4.4 & 0.0 & 0.0 \\
 & ASearcher-7B
 & 16.5 & 26.0 & 2.8 & 5.8 & 0.0 & 0.0 \\
 & DeepSeek-R1-671B
 & 41.3 & 55.1 & 20.7 & 31.7 & 0.4 & 1.5 \\
\midrule

\rowcolor{lightred} \cellcolor{white}
\multirow{5}{*}{\textit{\shortstack{Multi-Agent\\System}}} 
 & \ours-4B
 & 40.0 & 51.8  & 15.3 & 24.4 & 0.4 & 1.0 \\
 & Qwen3-4B
 & 31.2 & 42.3 & 8.4 & 15.5 & 0.0  & 0.0 \\
 & AgentFlow-7B
 & 28.7 & 45.4 & 9.0 & 20.2 & 0.4 & 1.5  \\
 & OWL-8B
 & 20.2 & 29.3 & 3.1 & 5.8 & 0.0 & 0.0 \\
 & MiroFlow-8B
 & 23.7 & 37.7 & 5.8 & 12.7 & 0.4 & 1.0 \\
\bottomrule
\end{tabular}

    \label{tab:widesearch}
\end{table*}

To fully explore the potential of width scaling, \ours requires a substantial volume of broad information-seeking tasks to facilitate stable training via MARL.
However, two significant gaps persist in current open-source QA resources.
First, existing datasets~\cite{yang2018hotpotqa, trivedi2022musique, mialon2023gaia, wei2025browsecomp} are primarily tailored for depth scaling, prioritizing multi-hop reasoning directed at single-entity queries or short-form responses.
Second, while benchmarks for broad information seeking~\cite{wong2025widesearch, lan2025deepwidesearch} do exist, they are typically constrained by limited scale and heavy reliance on manual annotation, rendering them insufficient for data-intensive RL training.

To bridge these gaps, we develop a fully automated data construction pipeline to synthesize high-quality training instances consisting of schema-constrained queries and standardized tabular outputs.
As illustrated in Fig.~\ref{fig:data_pipeline}, our pipeline operates in three key stages:
(1) \textbf{Query Generation}, where we extract user intents from HybridQA~\cite{chen2020hybridqa} and refine them into complex, schema-constrained queries that mandate specific table structures and broad coverage;
(2) \textbf{Answer Generation}, where we prompt the model to generate two responses independently along with the unique column(s), enabling self-consistency verification; and
(3) \textbf{QA Pair Filtering}, where we rigorously screen the data by discarding instances with low consistency or insufficient difficulty, ensuring that only robust and challenging samples remain in the final dataset.
We elaborate on the implementation details of each stage in the subsequent sections. 
Further analysis and detailed statistics of the constructed dataset are provided in Appendix~\ref{appendix:dataset}.

\subsection{Query Generation}
This stage extracts raw user intents and transforms them into complex, schema-constrained queries that mandate broad information coverage.
We utilize HybridQA~\cite{chen2020hybridqa} as our seed corpus due to its extensive scale and broad topical coverage derived from Wikipedia.
First, we extract the underlying user intent from the source data and stochastically sample a target row count between 10 and 50. This variation is explicitly introduced to enhance the diversity of the training data and ensure a broad retrieval scope. 
Second, we synthesize an initial query conditioned on these inputs.
Third, we refine the query with strict constraints, including standardized formats and column definitions. This minimizes ambiguity to facilitate the generation of a consistent and unique ground truth table for the subsequent stage.

\subsection{Answer Generation}
Taking the refined query as input, this stage generates candidate tabular responses and structural identifiers to facilitate quality verification.
First, we leverage Gemini to synthesize two independent responses for each refined query. This redundancy enables consistency filtering in Stage 3 to assess the quality of the answers.
Second, we instruct the model to identify the "unique column(s)" defined as the minimal set of column names required to distinguish one row from another. This identifier facilitates robust alignment, enabling the accurate mapping of predicted rows to the ground truth regardless of row permutations or discrepancies.

\subsection{QA Pair Filtering}
To guarantee the reliability of the synthesized data, in this stage, we implement a rigorous two-step filtering mechanism. 
First, we evaluate factual consistency by performing a cell-wise comparison between the two independent responses generated in Stage 2. Instances falling below a strict threshold 0.9 are discarded to eliminate ambiguous queries or model hallucinations.
Second, we apply a complexity filter that removes simplistic results, such as tables with fewer than 3 rows, to maintain sufficient difficulty.
Only samples passing both criteria are retained, resulting in a high-quality final dataset comprising the refined query, the canonical answer, and the unique column(s).

\section{Experiments}
\label{sec:exp}

To explore width scaling and demonstrate the effectiveness of \ours, we conduct experiments from four perspectives: main results on the WideSearch benchmark, width scaling behavior, ablation studies, and performance on standard QA benchmarks.

\textbf{Setup.}
We train \ours-4B from Qwen3-4B~\cite{yang2025qwen3} in thinking mode on a hybrid dataset that combines our constructed data with the standard QA data from ASearcher~\cite{gao2025beyond} in equal proportions. During training, the lead agent is allowed to invoke up to 10 parallel subagents per turn, with a maximum of 10 turns for the lead agent and 20 turns for each subagent. To improve training efficiency and reduce cost, we use an offline local knowledge base constructed from Wiki2018~\cite{karpukhin2020dense} as the \texttt{search} and \texttt{access} tools.
To ensure a fair comparison, we further train a single-agent variant, SingleSeek-R1-4B, using the same data and tools and increasing the maximum number of turns to 50 to allow sufficient tool interaction.
More training details are provided in Appendix~\ref{appendix:training}.

\textbf{Baselines.}
We compare against strong single-agent and multi-agent baselines. Single-agent baselines include Search-R1 (7B)~\cite{jin2025search} and ASearcher (7B)~\cite{gao2025beyond}. Multi-agent baselines include AgentFlow (7B)~\cite{li2025flow}, OWL~\cite{hu2025owl}, and MiroFlow~\cite{2025mirothinker}. Since MiroFlow is a framework and OWL only releases a 32B model, we implement their workflows with Qwen3-8B without additional training. In addition, we evaluate the base Qwen3-4B in a multi-agent system to demonstrate the gains from MARL training.
More evaluation details are provided in Appendix~\ref{appendix:evaluation}.

\subsection{Main Results}
\label{sec:exp:main_results}

We evaluate \ours-4B on the WideSearch~\cite{wong2025widesearch} benchmark to show the effectiveness of our multi-agent system trained via MARL for broad information seeking. 
The benchmark consists of 200 tasks, with 100 English and 100 Chinese queries requiring tabular output.
We report item F1 score, row F1 score, and Success Rate (SR). Each task is sampled four times, and we report Avg@4 for all metrics, Max@4 for F1 scores, and Pass@4 for SR.

As shown in Table~\ref{tab:widesearch}, \ours-4B achieves the best results on five out of six metrics among 4B and 8B baselines. The multi-agent system consistently outperforms the single-agent variant, yielding an absolute improvement of 11.9\% in item F1 score.
When compared to the base Qwen3-4B in the same multi-agent setting, our model achieves an 8.8\% improvement in item F1 score, showing that MARL training unlocks the potential of multi-agent systems.

Notably, \ours-4B attains performance comparable to DeepSeek-R1-671B in the single-agent setting, despite using nearly 170$\times$ fewer parameters.

\subsection{Exploring Width Scaling}
\label{sec:exp:wide_scaling}

\begin{figure}[t]
\centering
\includegraphics[width=\linewidth]{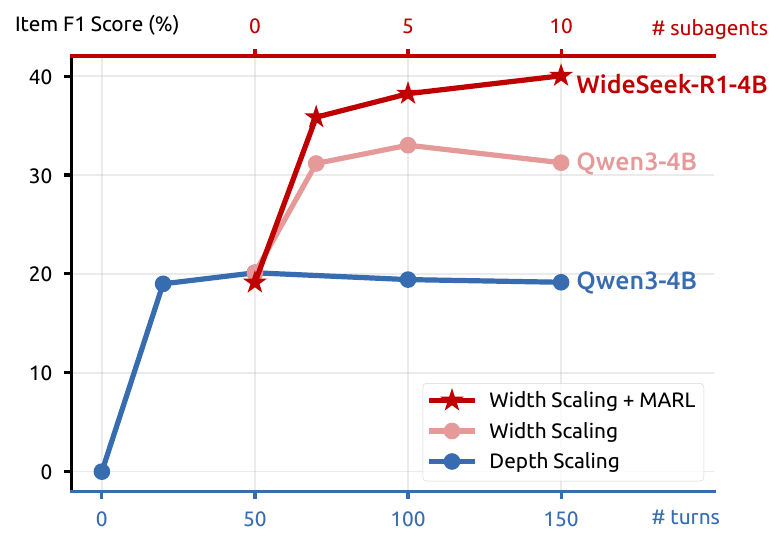}
\caption{
Comparison of depth and width scaling in performance with respect to (w.r.t.) test-time compute. The blue curve shows depth scaling in performance w.r.t. the number of turns (bottom axis), while the two red curves show width scaling in performance w.r.t. the number of subagents (top axis).
}
\label{fig:exp_scaling}
\end{figure}

To compare with depth scaling and illustrate the width scaling property of \ours, we plot the performance curves with respect to (w.r.t) test-time compute.
Specifically, for depth scaling, we consider the single-agent setting and plot a blue performance curve of Qwen3-4B w.r.t. the number of turns.
For width scaling, we adopt a multi-agent system with a fixed number of turns and plot two red performance curves of Qwen3-4B and \ours-4B w.r.t. the number of parallel subagents in one turn.

As shown in Fig.~\ref{fig:exp_scaling}, the best performance is achieved by \ours-4B with width scaling via MARL training.
Under depth scaling, the base model rapidly saturates. 
While additional turns initially yield gains, the performance quickly plateaus as the single agent is bottlenecked by its fixed context length.
Once depth scaling plateaus, we fix the number of turns and switch to width scaling by increasing the number of parallel subagents. 
For the base model, while width scaling initially yields improvements, its performance begins to decline when the number of subagents increases to ten. This deterioration is likely due to the accumulation of noise from conflicting responses, which overwhelms the untrained lead agent's ability to aggregate information effectively.
In contrast, \ours-4B demonstrates consistent performance gains with the number of subagents and pushes the frontier of width scaling to 40\% item F1 score with 10 subagents.
This demonstrates that \ours-4B unlocks the potential of width scaling via MARL by jointly optimizing the entire system: the lead agent masters orchestration, while sub-agents learn to deliver higher-quality responses in parallel. This synergy allows \ours to effectively harness the computational power of multi-agent system.

\subsection{Ablation Studies}
\label{sec:exp:ablation}

In this section, we conduct ablation studies to dissect the key components of our framework. We aim to answer two primary questions: (1) Is the joint optimization of both the lead agent and subagents necessary for optimal performance? (2) How does our constructed dataset impact the model's overall capability?

\begin{figure}[t]
\centering
\includegraphics[width=0.95\linewidth, trim=0 10 0 0, clip]{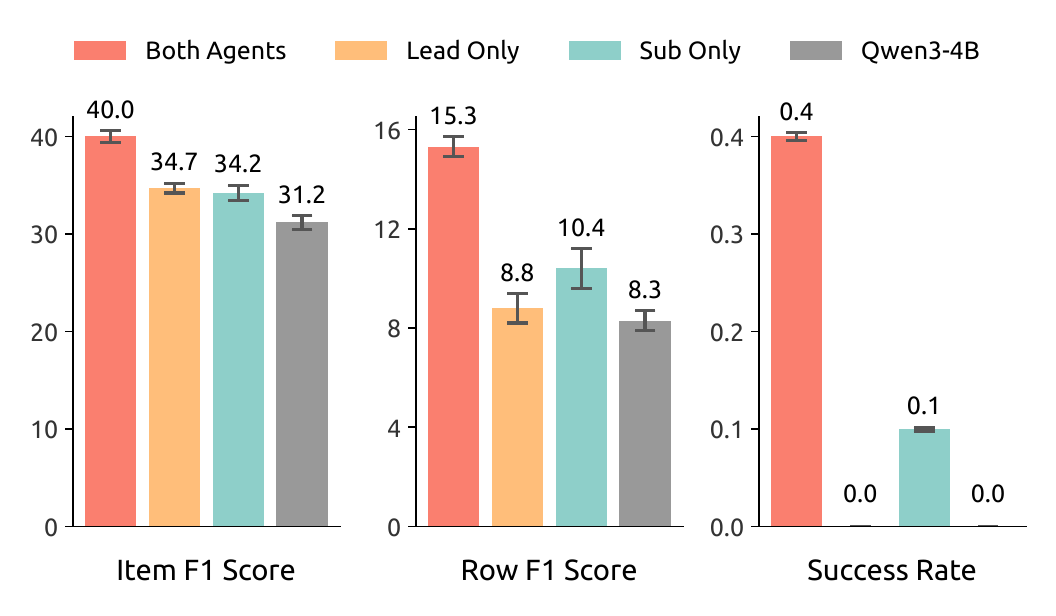}
\caption{Ablation study on lead agent and subagents by assigning \ours-4B to different roles.}
\label{fig:ablation_lead_and_sub}
\end{figure}

\begin{figure}[t]
\centering
\includegraphics[width=0.95\linewidth, trim=0 10 0 0, clip]{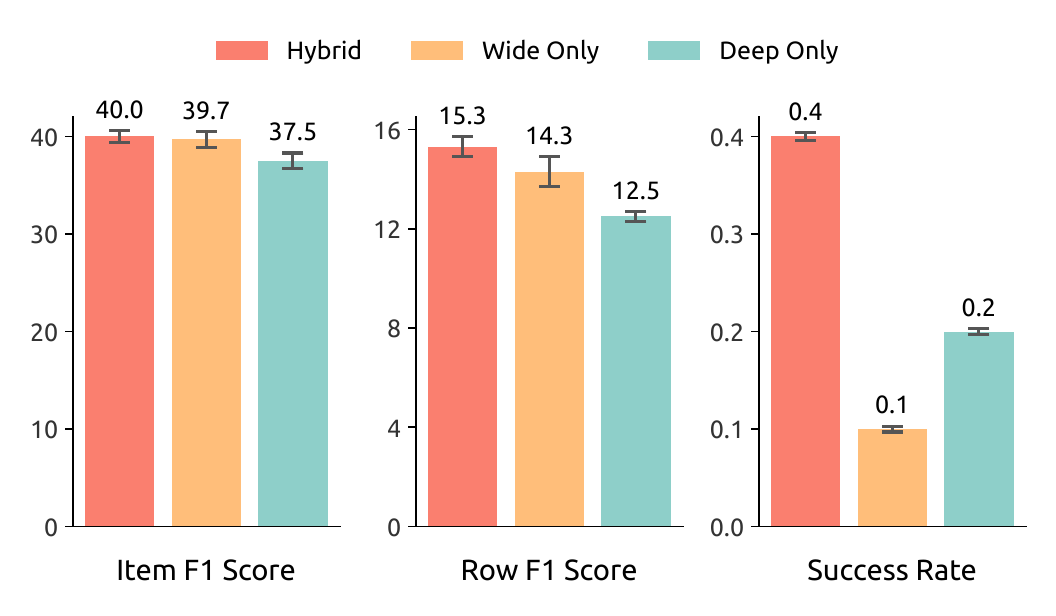}
\caption{Ablation study on training data by comparing models trained on hybrid, wide-only, and deep-only datasets. }
\label{fig:ablation_data}
\end{figure}

\begin{table*}[t]
    \centering
    \caption{
    Results on Single-Hop and Multi-Hop QA. 
    We report Avg@4 scores across seven widely used single-hop and multi-hop benchmarks. Notably, \ours-4B surpasses larger multi-agent systems like OWL-8B and MiroFlow-8B. These results validate that our MARL framework effectively enhances width scaling without compromising general reasoning capabilities.    
    }
    \setlength{\tabcolsep}{5pt}
\begin{tabular}{cccccccccc}
\toprule
\multirow{2}{*}{\textbf{Setting}} & \multirow{2}{*}{\textbf{Model}}
& \multirow{2}{*}{\textbf{Average}} & \multicolumn{3}{c}{\textbf{Single-Hop}}
& \multicolumn{4}{c}{\textbf{Multi-Hop}} \\
\cmidrule(lr){4-6} \cmidrule(lr){7-10}
& & & NQ & TriviaQA & PopQA
  & 2Wiki & HotpotQA & Bamboogle & MuSiQue \\
\midrule

\multirow{4}{*}{\textit{\shortstack{Single\\Agent}}}
 & SingleSeek-R1-4B
 & 57.0 & 58.8 & 78.3 & 48.0 & 70.9 & 62.1 & 54.6 & 26.5 \\
 & Qwen3-4B
 & 48.3 & 48.5 & 68.7 & 43.0 & 58.9 & 51.4 & 48.2 & 19.2 \\
 & Search-R1-7B
 & 55.4 & 49.9 & 78.0 & 55.7 & 58.1 & 60.8 & 58.4 & 27.1 \\
 & ASearcher-7B
 & 61.0 & 54.5 & 79.3 & 55.9 & 77.6 & 67.6 & 60.0 & 32.6 \\
\midrule

\rowcolor{lightred} \cellcolor{white}
& \ours-4B
& 59.0 & 56.1 & 78.5 & 48.5 & 75.0 & 64.2 & 61.8 & 28.9 \\

& Qwen3-4B
& 51.3 & 49.6 & 70.7 & 44.9 & 65.0 & 54.3 & 52.6 & 21.7 \\

& AgentFlow-7B
& 61.0 & 58.5 & 87.0 & 52.5 & 77.2 & 57.0 & 69.6 & 25.3 \\

& OWL-8B
& 57.2 & 64.0 & 74.2 & 52.2 & 62.6 & 61.0 & 55.8 & 30.4 \\

\multirow{-5}{*}{\textit{\shortstack{Multi-Agent\\System}}} 
& MiroFlow-8B
& 50.0 & 50.9 & 73.1 & 42.8 & 58.6 & 52.4 & 50.8 & 21.3 \\

\bottomrule
\end{tabular}

    \label{tab:qa}
\end{table*}

\textbf{Lead Agent and Subagents.}
To examine the individual impact of the lead agent and subagents, we perform an ablation study by varying the underlying model for each role. Specifically, we evaluate four settings by assigning either \ours-4B or Qwen3-4B to the lead agent and subagents, respectively, covering all possible combinations.

As shown in Fig.~\ref{fig:ablation_lead_and_sub}, the best performance is achieved when both the lead agent and subagents use \ours-4B. 
Regarding the item F1 score, upgrading either the lead agent or the subagents yields comparable gains over the base model. This suggests that our training effectively enhances both the orchestration capabilities of the lead agent and the subtask execution proficiency of the subagents. Crucially, the further gains observed when combining both roles highlight the synergy between these capabilities.

Notably, assigning \ours-4B to the subagents leads to a higher row F1 score and SR than assigning it solely to the lead agent. 
This disparity further underscores the substantial improvements in subtask execution achieved via our training. Since subagents are directly responsible for interacting with tools and solving specific queries, their enhanced execution capability is pivotal for meeting the strict criteria of row-level accuracy and overall task success.

These results indicate that both the lead agent and subagents are critical to the multi-agent system, and their joint optimization is necessary for optimal performance, validating the importance of end-to-end training of the entire system.

\textbf{Training Data.}
To demonstrate the effectiveness of our training data, we conduct an ablation study by training \ours-4B on different datasets: a wide-only dataset containing broad information-seeking data, a deep-only dataset consisting solely of training data from ASearcher, and a hybrid dataset that combines both in equal proportions. 
For a fair comparison, we ensure that the total number of training samples is identical across all three settings, and all other training parameters are kept constant.

As shown in Fig.~\ref{fig:ablation_data}, the model trained on the hybrid dataset consistently outperforms those trained on either wide-only or deep-only data across item F1 score, row F1 score, and Success Rate (SR). This result indicates that wide and deep data provide complementary benefits: wide data helps the system learn effective orchestration across parallel subagents, while deep data enhances information-seeking and subtask-solving capabilities. By combining both types of data, \ours-4B achieves the best performance.

\subsection{Standard QA Benchmarks}
\label{sec:exp:QA}

To assess the versatility of our method beyond broad information seeking, we further evaluate \ours on standard open-domain QA benchmarks. Our evaluation suite encompasses three single-hop datasets: Natural Questions (NQ)~\cite{kwiatkowski2019natural}, TriviaQA~\cite{joshi2017triviaqa}, and PopQA~\cite{mallen2023not}, and four multi-hop datasets: 2WikiMultiHopQA~\cite{ho2020constructing}, HotpotQA~\cite{yang2018hotpotqa}, Bamboogle~\cite{press2023measuring}, and MuSiQue~\cite{trivedi2022musique}.
To measure the robustness of our system against both single-agent and multi-agent baselines, we report the Avg@4 assessed via an LLM-as-a-judge evaluation across these diverse tasks.

As presented in Table~\ref{tab:qa}, \ours-4B achieves a superior average score of 59.0\%. Compared to its direct backbone, multi-agent Qwen3-4B, our method achieves a significant gain of 7.7\%, demonstrating the effectiveness of our MARL training. Moreover, \ours-4B also surpasses our trained single-agent variant, SingleSeek-R1-4B, by 2.0\%, indicating that our multi-agent framework yields consistent benefits even on standard QA tasks beyond broad information seeking.
Furthermore, despite its compact 4B size, \ours surpasses larger multi-agent systems like OWL-8B and MiroFlow-8B. 
These results validate that our MARL framework effectively enhances width scaling without compromising general reasoning capabilities.

\section{Conclusion}
\label{sec:conclusion}
In this work, we explored width scaling as a complementary dimension to the prevailing depth scaling paradigm in LLMs. We proposed \ours, a multi-agent framework trained via MARL that synergizes scalable orchestration with parallel execution to tackle broad information-seeking tasks. By shifting the focus from individual competence to organizational capability, \ours-4B achieves performance comparable to single-agent DeepSeek-R1-671B on the WideSearch benchmark. Crucially, our experiments demonstrate that while depth scaling faces diminishing returns, width scaling exhibits consistent performance gains as the number of parallel subagents increases. Furthermore, the release of our curated 20k dataset provides a foundation for future research in training scalable multi-agent systems. We hope this work inspires further exploration into efficient, parallelized agent organizations that can solve complex problems beyond the reach of individual LLM agents.

\section*{Acknowledgements}
\label{sec:ack}
We thank the ASearcher team for open-sourcing high-quality multi-hop QA datasets~\cite{gao2025beyond}. We thank Tianchen Zhao and Tianyu Fu for constructive discussions.

This work was supported by the National Natural Science Foundation of China (No.62406159, 62325405), Ant Group, Beijing National Research Center for Information Science, Technology (BNRist), and Beijing Innovation Center for Future Chips.

\section*{Impact Statement}
\label{sec:impact}

We introduce \ours, a framework that leverages width scaling to achieve high-performance broad information seeking using significantly smaller models with 4B parameters compared to existing state-of-the-art solutions.

\textbf{Exploring LLM Scaling Dimension.} While recent trends prioritize depth scaling, our work highlights the potential of width scaling via multi-agent collaboration. By validating that parallelized small agents can rival giant single models, we encourage the community to explore efficient collaborative architectures beyond simple parameter scaling.

\textbf{Democratization of AI.} Achieving performance comparable to massive models exceeding 600B parameters while using only 4B parameters significantly lowers computational barriers. This democratizes access to advanced reasoning capabilities, enabling researchers and organizations with limited compute resources to deploy high-performance systems on modest hardware.

\textbf{Potential Risks.} We acknowledge that autonomous agent swarms could be misused for scalable automated data gathering or misinformation generation. This underscores the necessity of developing robust safety guardrails and usage policies for responsible real-world deployment.

\bibliography{reference}
\bibliographystyle{icml2026}

\newpage
\appendix
\onecolumn

\section{Limitation}
\label{appendix:limitation}
\paragraph{Model Size.}
Currently, we only evaluate the Qwen3-4B model in the training setting, as training a reasoning-based multi-agent system is computationally expensive. Even with a 4B-parameter model, training requires approximately 3,000 GPU hours on H100 GPUs, resulting in substantial computational cost.

\paragraph{Credit Assignment.}
In our current MARL framework, the reward signal is derived from the final task outcome and shared across the agent group. This introduces a challenge in structural credit assignment: distinguishing whether a failure stems from the lead agent's flawed orchestration or a subagent's execution error. While our training stabilizes overall performance, this coarse-grained feedback may limit the system's ability to correct specific sub-optimal behaviors within the hierarchical chain. Future work could explore more granular, role-specific reward modeling to address this ambiguity.

\paragraph{Nested Hierarchy.} 
Although our underlying framework supports flexible, recursive agent workflows, we intentionally restrict the system to a static two-layer hierarchy during training to ensure optimization stability. Specifically, we disable the capability for subagents to recursively spawn their own subagents (i.e., a "main-sub-sub" structure). Allowing unbounded recursive delegation introduces variable trajectory structures and explodes the state space, which we found to severely destabilize the MARL training process. Consequently, while this constraint guarantees convergence, it limits the system's ability to autonomously deepen its organizational structure for unexpectedly complex sub-tasks.

\paragraph{Training Efficiency.} 
Profiling shows that nearly 90\% of the training step time is dominated by rollout, primarily due to long-tail generation. This overhead stems from our use of collocated RL training for improved stability, where the training process must wait for rollout generation to complete before proceeding. 
While this design ensures stable optimization, it significantly increases training latency. In future work, we plan to explore more efficient training paradigms, such as asynchronous rollout or decoupled generation and training, to improve training efficiency.

\section{Rollout Detail}
\label{appendix:rollout}

In this section, we claim the details of \ours's multiagent system

\subsection{System Design}
Given a query $q$ from dataset $\mathcal{D}$, we generate $G$ rollouts $\{\tau_i\}_{i=1}^{G}$. Rollout $\tau_i$ contains $N_i$ agent trajectories,
\[
\tau_i = \{\tau_{i,1}, \tau_{i,2}, \dots, \tau_{i,N_i}\},
\]
where agent $a=1$ is the lead agent, and agents $a \in \{2,\dots,N_i\}$ are subagents. Each agent trajectory is multi-turn, with $T_{i,a}$ turns.

For the lead agent ($a=1$), we write the state at turn $t$ as
\[
s_{i,1}^{t}
=
\big[p_{\mathrm{lead}},\; q,\; o_{i,1}^{1},\; tcr_{i,1}^{1},\; \dots,\; o_{i,1}^{t-1},\; tcr_{i,1}^{t-1}\big].
\]
For a subagent ($a \ge 2$), we write
\[
s_{i,a}^{t}
=
\big[p_{\mathrm{sub}},\; q_a,\; o_{i,a}^{1},\; tcr_{i,a}^{1},\; \dots,\; o_{i,a}^{t-1},\; tcr_{i,a}^{t-1}\big].
\]
Here, $p_{\mathrm{lead}}$ and $p_{\mathrm{sub}}$ are the system prompts for the lead agent and subagents. $q$ is the original task, and $q_a$ is the subtask delegated by the lead agent to subagent $a$.

At turn $t$, agent $a$ generates a token sequence
\[
o_{i,a}^{t} = \big[o_{i,a}^{t,1}, o_{i,a}^{t,2}, \dots \big],
\]
where token $o_{i,a}^{t,j}$ is sampled from
\[
\pi_\theta\!\left(o_{i,a}^{t,j}\mid s_{i,a}^{t},\, o_{i,a}^{t,<j}\right).
\]
We denote the extracted tool call as
\[
tc_{i,a}^{t} = \mathrm{Extract}_{\mathrm{tool}}\!\left(o_{i,a}^{t}\right),
\qquad
tcr_{i,a}^{t} = \mathrm{Tool}\!\left(tc_{i,a}^{t}\right),
\]
where $tcr_{i,a}^{t}$ is the tool result returned to the agent.

Suppose that at some turn $t$, the lead agent spawns a set of subagents, and these subagents finish their last turns. To keep the lead agent context short, we construct the lead agent tool context by collecting the final-turn outputs of these subagents and removing the thinking content:
\[
tc_{i,1}^{t}
=
\mathrm{Discard}_{\mathrm{think}}\!\left(
o_{i,a_1}^{T_{i,a_1}}, \dots, o_{i,a_m}^{T_{i,a_m}}
\right).
\]
For subagents, tool calls are obtained directly from search and browser actions in their own outputs.

\subsection{Reward Design}
As discussed in Sec.~\ref{sec:method:MARL}, we assign an outcome reward $R_i$ to each rollout $\tau_i$ and compute a group normalized advantage $\hat{A}_i$ across $G$ rollouts. All tokens within rollout $i$ share the same advantage $\hat{A}_i$.

The reward is defined as
\begin{equation}
R_i
=
\begin{cases}
r_{\mathrm{ans}} + r_{\mathrm{format}} + r_{\mathrm{tool}} - r_{\mathrm{len}},
& \text{valid format}, \\[4pt]
0,
& \text{invalid format}.
\end{cases}
\end{equation}

where

\begin{itemize}
    \item $r_{\mathrm{ans}}$: the Item F1 score of the generated answer compared with the ground truth.
    \item $r_{\mathrm{format}}$: a binary reward indicating whether the generated answer follows a valid Markdown structure.
    \item $r_{\mathrm{tool}}$: a binary reward indicating whether the rollout invokes a browser tool at least once.
    \item $r_{\mathrm{len}}$: a length penalty designed to discourage excessively long responses. 
    When the response length $L$ exceeds a threshold $L_\mathrm{thr}$, we apply a linearly scaled penalty. The penalty is clipped at a hard length limit $L_{\max}$, where $\alpha_{\mathrm{len}}$ controls the penalty strength.
\end{itemize}

The length penalty is computed as
\begin{equation}
r_{\mathrm{len}}
=
\begin{cases}
\alpha_{\mathrm{len}} \cdot
\operatorname{clip}\!\left(
\dfrac{L - L_\mathrm{thr}}
{L_{\max} - L_\mathrm{thr}},
0, 1
\right),
& L > L_\mathrm{thr}, \\[6pt]
0, & L \le L_\mathrm{thr}.
\end{cases}
\end{equation}

\subsection{Collection Buffer}
We apply additional filtering rules when constructing the training buffer.

\begin{itemize}
    \item If the final answer format is valid, we only add trajectories that stay within the maximum context limit and the maximum allowed number of turns. This prevents assigning positive rewards to subagents that fail, even when the lead agent produces a correct final answer.

    \item If the final answer format is invalid, we always add the final turn of the lead agent to penalize formatting errors. In addition, if any turn exhibits clear repeated token loops that lead to context overflow, we add those turns to penalize repetition. Otherwise, for trajectories that exceed the maximum context limit or the maximum allowed number of turns, we add all turns in those trajectories so that the model learns to avoid these failure modes.
\end{itemize}

\section{Dataset Detail}
\label{appendix:dataset}
\subsection{Dataset Statistics}

In this section, we provide details regarding the synthetic dataset used in our experiments.
Our constructed dataset consists of 20,000 instances.
Fig.~\ref{fig:data_distribution} visualizes the structural distribution of the constructed dataset. As illustrated, the row counts exhibit a broad spectrum with a median of 30, effectively capturing varying degrees of retrieval complexity, while the column counts remain tightly clustered around a median of 6. This structural diversity ensures that the dataset serves as a robust testbed for evaluating agent performance across different problem scales.

During the construction phase, we observed that \texttt{GPT-4} and \texttt{Claude-4-sonnet} exhibited suboptimal performance in generating high-quality queries and corresponding ground truth for this specific task. Consequently, we utilized \texttt{gemini-3-pro-preview} for data synthesis. The generation cost was approximately \$0.10 per instance. The final dataset reflects a retention rate of 73.28\%, demonstrating that our Stage 3 filtering effectively balances strict quality control with cost-effectiveness. 

\begin{figure}[htbp] %
  \centering
  \hfill
  \begin{subfigure}{0.45\textwidth}
    \centering
    \includegraphics[width=\textwidth]{figs/data_row_count.pdf} 
    \caption{Row Count Distribution}
    \label{fig:data_row_distribution}
  \end{subfigure}
  \hfill
  \begin{subfigure}{0.45\textwidth}
    \centering
    \includegraphics[width=\textwidth]{figs/data_column_count.pdf}
    \caption{Column Count Distribution}
    \label{fig:data_column_distribution}
  \end{subfigure}
  \hfill
  
  \caption{
  Statistical distribution of ground truth answer table dimensions in the constructed dataset (N=20,000).
  }
  \label{fig:data_distribution}
\end{figure}

\subsection{Sample Instance}
\label{app:sample_instance}

To provide a concrete illustration of the synthesized data, we present a representative sample instance in its raw JSON format. This instance, as shown in Fig.~\ref{fig:sample_instance}, exemplifies the structural complexity and the alignment between the query and the ground truth that our pipeline generates. 

Each instance in our dataset is formatted as a JSON object containing three key fields:
\begin{itemize}
    \item \texttt{question}: The natural language query generated by our construction pipeline. It explicitly integrates formatting instructions alongside the retrieval task, mandating that the output be a table with specific column definitions and data presentation styles. 
    \item \texttt{answer}: The ground truth response generated by our construction pipeline, formatted as a Markdown table that perfectly satisfies the constraints in the question.
    \item \texttt{unique\_columns}: A set of column names used to uniquely distinguish rows. This identifier is critical for robust alignment: it allows the evaluation metric to accurately map predicted rows to the ground truth regardless of row permutations or row-wise discrepancies.
\end{itemize}

\subsection{Evaluation Metrics}
To comprehensively evaluate model performance on our dataset, we utilize three metrics that assess generation quality across progressive levels of granularity ranging from fine-grained cell accuracy to holistic table correctness:
\begin{itemize}
    \item \textbf{Item F1 Score}: Treats each cell as one unit and computes the F1 score by comparing the predicted cells against the ground truth.
    \item \textbf{Row F1 Score}: Treats each row as one unit. A predicted row is deemed correct only if it matches a ground-truth row. The final score is calculated as the F1 measure over the set of rows. 
    \item \textbf{Success Rate}: Represents the proportion of queries where the generated table perfectly matches the ground truth, indicating a complete and accurate retrieval.
\end{itemize}

\begin{figure}[htbp]
\centering
\definecolor{guide_green}{RGB}{229, 244, 229}
\definecolor{guide_green_frame}{RGB}{127, 201, 127}
\definecolor{guide_orange}{RGB}{255, 242, 228}
\definecolor{guide_orange_frame}{RGB}{255, 144, 27}
\definecolor{guide_purple}{RGB}{229, 223, 237}
\definecolor{guide_purple_frame}{RGB}{140, 112, 177}

\begin{tcolorbox}[
    colback=guide_green,
    colframe=guide_green_frame,
    title=\texttt{question},
    arc=3mm
]
\begin{small}
I am conducting research on the conservation geography of New Zealand and need a structured overview of its National Parks system. I need you to identify all National Parks in New Zealand that were active and designated as of December 31, 2017, excluding any parks disestablished before that date, and compile their details. Please output the organized data as a single Markdown table, do not split into multiple markdown tables, each cell must be filled according to the column requirements, no omissions allowed, output in English.

The column names are as follows:
National Park, Establish Year, Total Area (km2), Primary Island, Administering Regional Councils

Do not ask me any questions, just output the result in the format:
\begin{verbatim}
```markdown
{data_content}
```
\end{verbatim}
Output only the table header and rows; do not add analysis, commentary, or any additional text.
\end{small}
\end{tcolorbox}

\begin{tcolorbox}[
    colback=guide_orange,
    colframe=guide_orange_frame,
    title=\texttt{answer},
    arc=3mm
]
\begin{small}
\begin{center}
\resizebox{\textwidth}{!}{%
\begin{tabular}{l l l l l}
\toprule
National Park & Establish Year & Total Area (km2) & Primary Island & Administering Regional Councils \\
\midrule
Tongariro National Park & 1887 & 786 & North Island & Manawatū-Whanganui \\
Egmont National Park & 1900 & 342 & North Island & Taranaki \\
Arthur's Pass National Park & 1929 & 1,185 & South Island & Canterbury, West Coast \\
Abel Tasman National Park & 1942 & 237 & South Island & Tasman \\
Fiordland National Park & 1952 & 12,607 & South Island & Southland \\
Aoraki/Mount Cook National Park & 1953 & 707 & South Island & Canterbury \\
Nelson Lakes National Park & 1956 & 1,019 & South Island & Tasman \\
Westland Tai Poutini National Park & 1960 & 1,320 & South Island & West Coast \\
Mount Aspiring National Park & 1964 & 3,562 & South Island & Otago, West Coast \\
Whanganui National Park & 1986 & 742 & North Island & Manawatū-Whanganui \\
Paparoa National Park & 1987 & 430 & South Island & West Coast \\
Kahurangi National Park & 1996 & 4,520 & South Island & Tasman, West Coast \\
Rakiura National Park & 2002 & 1,400 & Stewart Island & Southland \\
\bottomrule
\end{tabular}%
}
\end{center}
\end{small}
\end{tcolorbox}

\begin{tcolorbox}[
    colback=guide_purple,
    colframe=guide_purple_frame,
    title=\texttt{unique\_columns},
    arc=3mm
]
\begin{small}
["National Park"]
\end{small}
\end{tcolorbox}

\caption{A representative JSON instance from the synthesized dataset, including the question, answer, and unique columns.}
\label{fig:sample_instance}
\end{figure}

\section{Training Detail}
\label{appendix:training}
All experiments were conducted on the Qwen3-4B model with thinking mode enabled. Training was performed on NVIDIA H100 GPUs: \ours-4B was trained using 32 H100s, while SingleSeek-R1-4B was trained using 16 H100s. For all runs, the batch size was fixed at 128, and the maximum context length was set to 32K to better support long chain-of-thought reasoning. Each experiment was trained for a total of 150 steps.

To improve training efficiency, we used RLinf\footnote{\url{https://github.com/RLinf/RLinf}}, a flexible and scalable reinforcement learning infrastructure. For efficient rollouts, we adopted SGLang with a tensor parallel size of 1 and a GPU memory utilization ratio of 0.5. The sampling parameters were set to a temperature of 1.0 and top-$p$ of 1.0. For efficient optimization, we used Megatron with a tensor parallel size of 2 and a constant learning rate of $1\times 10^{-6}$. \(\epsilon_{\text{low}}\) and \(\epsilon_{\text{high}}\) are set to 0.2 and 0.28, respectively. For reward design, we set $r_{\mathrm{format}}=0.1$, $r_{\mathrm{tool}}=0.05$, $\alpha_{\mathrm{len}}=0.1$, $L=3000$, and $L_{\max}^{\mathrm{hard}}=5000$.

For \ours-4B, the maximum number of parallel sub-agents that can be invoked per turn was set to 10. The maximum number of turns was 10 for the lead agent and 20 for each subagent. In addition, the maximum number of parallel \texttt{search} and \texttt{access} tool calls per turn was set to 5. For SingleSeek-R1-4B, the total number of allowed turns was set to 50, and the model was restricted to calling at most one tool per turn.

Finally, for hybrid-dataset training, we organized the mixed data stream such that every batch contained 64 deep samples and 64 wide samples. This batching strategy helps keep the loss and gradients smooth across batches, thereby improving training stability.

\section{Evaluation Detail}
\label{appendix:evaluation}
\label{appendix:evaluation_details}

In this section, we detail the implementation of baseline methods compared in our experiments. 
We benchmark \ours against both representative single-agent systems and state-of-the-art multi-agent frameworks. 

\subsection{Benchmark}

\paragraph{Widesearch.}
Widesearch~\cite{wong2025widesearch} is a comprehensive benchmark designed to evaluate search agents. Distinct from standard QA benchmarks that typically target specific answer retrieval, Widesearch emphasizes \textbf{broad information seeking}, explicitly requiring agents to collect and synthesize attributes of multiple entities and output the final response in a structured tabular format. The benchmark consists of a total of 200 questions, balanced with 100 English and 100 Chinese queries.
We employ the official evaluation code provided by the Widesearch benchmark. Following the standard protocol, we use \texttt{gpt-4.1-2025-04-14} as the LLM judge.
We report Item F1, Row F1, and Success Rate. For Item F1 and Row F1, we present both Avg@4 and Max@4. For Success Rate, we report Avg@4 and Pass@4.

\paragraph{Standard QA.}
For standard QA tasks, we evaluate on a suite of 7 datasets, categorized into two groups based on reasoning complexity. The single-hop group includes NQ~\cite{kwiatkowski2019natural}, TriviaQA~\cite{joshi2017triviaqa}, and PopQA~\cite{mallen2023not}. The multi-hop group comprises 2WikiMultiHopQA~\cite{ho2020constructing}, HotpotQA~\cite{yang2018hotpotqa}, Bamboogle~\cite{press2023measuring}, and MuSiQue~\cite{trivedi2022musique}.
We utilize the subsampled versions provided by Asearcher~\cite{gao2025beyond} rather than the full datasets. This is necessary because the original test sets contain a massive volume of samples, making evaluation computationally intensive for agentic workflows.
We report the performance using the Avg@4 metric.

\subsection{Single-Agent Baselines}
\label{app:single_agent_baselines}

\paragraph{SingleSeek-R1-4B}

\begin{itemize}
    \setlength\itemsep{0pt}
    \item \textbf{Model Configuration}: 
    SingleSeek-R1-4B is a single-agent variant trained on Qwen3-4B. It utilizes a hybrid dataset combining our broad information-seeking data (Sec.~4) with standard QA data from Asearcher~\cite{gao2025beyond} in a 1:1 ratio. The agent is equipped with both \texttt{search} and \texttt{access} tools. The total number of allowed turns is set to 50 to allow sufficient tool interaction, and the model is restricted to calling at most one tool per turn.%
    \item \textbf{WideSearch Evaluation}: The evaluation is conducted using the standard online toolset \texttt{Serper}\footnote{\label{fn:serper}\url{https://serper.dev/}} as \texttt{search} and \texttt{Jina}\footnote{\label{fn:jina}\url{https://jina.ai/}} as \texttt{access}.%
    
    \item \textbf{Standard QA Evaluation}: The evaluation is conducted using the standard offline toolset. The toolset comprising a local knowledge base constructed from Wiki2018~\cite{karpukhin2020dense} that serves as both \texttt{search} and \texttt{access} tools to simulate a controlled information environment.
\end{itemize}

\paragraph{Qwen3-4B}
\begin{itemize}
    \setlength\itemsep{0pt}
    \item \textbf{Model Configuration}: We deploy the Qwen3-4B model with the exact same setting as SingleSeek-R1-4B to ensure a fair comparison.
    \item \textbf{WideSearch Evaluation}: The evaluation is conducted using the standard online toolset \texttt{Serper}\textsuperscript{\ref{fn:serper}} as \texttt{search} and \texttt{Jina}\textsuperscript{\ref{fn:jina}} as \texttt{access}.%
    
    \item \textbf{Standard QA Evaluation}: The evaluation is conducted using the standard offline toolset.
\end{itemize}

\paragraph{Search-R1-7B}
\begin{itemize}
    \setlength\itemsep{0pt}
    \item \textbf{Model Configuration}: We use the released Search-R1-7B~\cite{jin2025search} model.
    \item \textbf{WideSearch Evaluation}: We configure the agent with only the \texttt{Serper}\textsuperscript{\ref{fn:serper}} as \texttt{search}. %
    The \texttt{access} tool is disabled to match the original training distribution of Search-R1 and avoid Out-Of-Distribution (OOD) performance degradation.

    \item \textbf{Standard QA Evaluation}: We directly report the results from Asearcher~\cite{gao2025beyond}.
\end{itemize}

\paragraph{Asearcher-7B}
\begin{itemize}
    \setlength\itemsep{0pt}
    \item \textbf{Model Configuration}: We utilize the Asearcher-7B~\cite{gao2025beyond} model, a specialized search agent.
    \item \textbf{WideSearch Evaluation}: The evaluation is conducted using the standard online toolset \texttt{Serper}\textsuperscript{\ref{fn:serper}} as \texttt{search} and \texttt{Jina}\textsuperscript{\ref{fn:jina}} as \texttt{access}.%
    \item \textbf{Standard QA Evaluation}: We directly report the results from Asearcher~\cite{gao2025beyond}.
\end{itemize}

\paragraph{DeepSeek-R1-671B}
\begin{itemize}
    \setlength\itemsep{0pt}
    \item \textbf{Model Configuration}: We consider the DeepSeek-R1-671B~\cite{guo2025deepseek} model as a strong baseline.
    \item \textbf{WideSearch Evaluation}: We directly report the results from Widesearch~\cite{wong2025widesearch}.
\end{itemize}

\subsection{Multi-Agent Baselines}
\label{app:multi_agent_baselines}

\paragraph{Qwen3-4B}
\begin{itemize}
    \setlength\itemsep{0pt}
    \item \textbf{Model Configuration}: We adapt the \ours framework using Qwen3-4B as the backbone for both planning and worker agents.
    \item \textbf{WideSearch Evaluation}: The evaluation is conducted using the standard online toolset \texttt{Serper}\textsuperscript{\ref{fn:serper}} as \texttt{search} and \texttt{Jina}\textsuperscript{\ref{fn:jina}} as \texttt{access}. %
    \item \textbf{Standard QA Evaluation}: The evaluation is conducted using the standard offline toolset.
\end{itemize}

\paragraph{AgentFlow-7B}
\begin{itemize}
    \setlength\itemsep{0pt}
    \item \textbf{Model Configuration}: We utilize the official \texttt{agentFlow-planner-7b} model, which is proposed from AgentFlow, as the leader agent. To maintain consistency with the original AgentFlow setup, we strictly follow their tool and subagent configuration, where the Google search tool is powered by Gemini-2.5-Flash.
    \item \textbf{WideSearch Evaluation}: We evaluate on the complete Widesearch dataset. The standard AgentFlow implementation invokes the Gemini-2.5-Flash, which contributes to relatively high evaluation scores on WideSearch, although it overall remains inferior to \ours-4B model. 
    \item \textbf{Standard QA Evaluation}: We independently evaluated NQ, TriviaQA, and PopQA. For the other four multi-hop datasets (2WikiMultiHopQA, HotpotQA, Bamboogle, MuSiQue), we directly report the results from AgentFlow~\cite{li2025flow}.
\end{itemize}

\paragraph{OWL-8B}
\begin{itemize}
    \setlength\itemsep{0pt}
    \item \textbf{Model Configuration}: We use OWL~\cite{hu2025owl} and replace the original GPT-series models with locally deployed Qwen3-8B for both Planner and Worker roles to ensure consistency.
    \item \textbf{WideSearch Evaluation}: The evaluation is conducted using the standard online toolset \texttt{Serper}\textsuperscript{\ref{fn:serper}} as \texttt{search} and \texttt{Jina}\textsuperscript{\ref{fn:jina}} as \texttt{access}. %
    \item \textbf{Standard QA Evaluation}: 
    The evaluation is conducted using the standard offline toolset.
\end{itemize}

\paragraph{MiroFlow-8B}
\begin{itemize}
    \setlength\itemsep{0pt}
    \item \textbf{Model Configuration}: We utilize MiroFlow~\cite{2025mirothinker} with Qwen3-8B as the backbone for all agents in the workflow.
    \item \textbf{WideSearch Evaluation}: The evaluation is conducted using the standard online toolset \texttt{Serper}\textsuperscript{\ref{fn:serper}} as \texttt{search} and \texttt{Jina}\textsuperscript{\ref{fn:jina}} as \texttt{access}. %
    \item \textbf{Standard QA Evaluation}: 
    The evaluation is conducted using the standard offline toolset.
\end{itemize}

\section{Pattern Analysis}
\label{appendix:pattern}
\begin{table*}[t]
    \centering
    \caption{Behavioral metrics after training in single-agent and multi-agent settings: average turns (total, lead, subagent), tool-call counts (\texttt{call\_subagent}, \texttt{search}, \texttt{access}), and answer format score.}
    \renewcommand{\arraystretch}{1.2}
\setlength{\tabcolsep}{5pt}

\begin{tabular}{ccccccccc}
\toprule
\multirow{2}{*}{\textbf{Setting}} &
\multirow{2}{*}{\textbf{Model}} &
\multicolumn{3}{c}{\textbf{Avg Turns}} &
\multicolumn{3}{c}{\textbf{Tool Call Count}} &
\multirow{2}{*}{\makecell{\textbf{Answer}\\\textbf{Format}}} \\
\cmidrule(lr){3-5}\cmidrule(lr){6-8}
& & Total & Lead Agent & Subagent
  & \texttt{call\_subagent} & \texttt{search} & \texttt{access} & \\
\midrule

\multirow{2}{*}{\textit{\shortstack{Single\\Agent}}}
 & SingleSeek-R1-4B
 & 7.0 &  - & - & - & 3.6 & 4.9 & 94.2\\
 & Qwen3-4B
 & 9.5 & - & - & - & 5.3 & 0.8 & 87.7\\
\midrule

\multirow{2}{*}{\textit{\shortstack{Multi-Agent\\System}}}
 & \ours-4B
 & 91.0 & 3.8 & 6.2 & 14.1 & 83.9 & 74.1 & 95.2 \\
 & Qwen3-4B
 & 23.2 & 2.3 & 2.9 & 7.2 & 23.3 & 11.9 & 97.1\\
\bottomrule
\end{tabular}

    \label{tab:pattern}
\end{table*}

In this section, we analyze how several key behavioral metrics change after training, as summarized in Table~\ref{tab:pattern}.

First, in the single-agent setting, SingleSeek-R1-4B improves the answer format score over the base model, but it still lower than the multi-agent system. We attribute this gap to the lack of context isolation: tool outputs are injected into a single shared context, which can introduce noise and formatting drift. In contrast, \ours-4B exhibits a slightly lower answer format score than the base model. Our profiling suggests that some queries are intrinsically difficult to resolve using an offline Wikipedia corpus; consequently, the model sometimes converges to a failure template (e.g., “I can’t solve this problem”), which hurts the format metric.

Second, multi-agent systems use substantially more turns per sample than single-agent baselines. Notably, \ours-4B produces nearly $4\times$ more total turns than the base model, with about $1.6\times$ more lead-agent turns and $2.1\times$ more subagent turns. This indicates that our MARL training encourages deeper interaction and more tool use, which in turn supports more confident answers.

Third, \ours-4B demonstrates improved width scaling by spawning roughly $2\times$ more subagents than the base model. Moreover, it increases the fraction of \texttt{access} calls from 33.8\% to 46.9\%. This is a desirable pattern: \texttt{search} snippets are often brief, and following the returned URLs via \texttt{access} allows the model to gather richer evidence before producing the final response.

\clearpage %

\section{Prompt Detail}
\label{appendix:prompts}

We present the system prompts used during the training phase of our framework. Specifically, we list the prompt for the lead agent and the subagents.

\subsection{Lead Agent System Prompt}
\definecolor{prompt_gray}{RGB}{245, 245, 245}
\definecolor{prompt_gray_frame}{RGB}{128, 128, 128}

\begin{tcolorbox}[
    colback=prompt_gray,
    colframe=prompt_gray_frame,
    title=\texttt{SYSTEM\_PROMPT\_LEAD\_AGENT},
    arc=3mm,
    breakable
]
\begin{small}
\begin{lstlisting}[breaklines=true, breakindent=0pt, basicstyle=\ttfamily\footnotesize, columns=fullflexible, numbers=none, frame=none]
# Role
You are a main-agent working on a hard task. Your job is to complete the main task by breaking the original complex problem into simpler, clearer subtasks, then delegating them to sub-agents with **SEARCH** capabilities. 

You must conduct reasoning inside <think> and </think> first every time you get new information.

# Tool Usage
After completing your reasoning, if you determine the main task is quite complex and requires additional knowledge, you may break the main question into smaller, more manageable **parallel** subtasks. You may delegate these subtasks to sub-agents using the **create_sub_agents** tool.

Keep in mind that sub-agents run **in parallel** and can search for information using additional tools. Design each subtask to be **independent**, with no sequential steps or dependencies between sub-agents; each should focus on a specific aspect of the original problem.

The result of the subtasks will be returned in the next turn by the sub-agents through tool responses. 

You can perform multiple turns of tool calls. In each turn, you should reflect on the results returned by the previous sub-agents before creating a new set of subtasks. Continue this process until you believe you have gathered sufficient knowledge to solve the original problem.

# Few-shot Examples

Below are two examples to guide you in better decomposing the original questions.

## First Example

**Question:**
Please help me compile a list of the top 10 individuals from China and the United States on the 2025 Forbes list. For each person, provide their name, Forbes ranking, country, birth year, and university attended (if not attended, fill in as "Nan").

**Your Approach:**
In the first turn, you should:

<think>
This question requires us to research the top 10 individuals from China and the U.S. on the 2025 Forbes list. To ensure accuracy, I must first identify who the top 10 individuals from each country are. Therefore, I will create two sub-agents with search capabilities: one to find the top 10 from China, and another to find the top 10 from the U.S. After that, I can proceed to gather more detailed information.
</think>

<tool_call>
{{"name": "create_sub_agents", "arguments": {{"sub_agents": [{{"prompt": "Find the top 10 individuals on the 2025 Forbes list from China and their rankings."}}, {{"prompt": "Find the top 10 individuals on the 2025 Forbes list from the U.S. and their rankings."}}]}}}}
</tool_call>

In the second turn, ideally, you will receive a complete list of 20 individuals (10 from each country) from the sub-agents. At this point, you should:

<think>
Based on the sub-agents' responses, I now know that the top 10 individuals from China are person1, person2, ..., person10, and from the U.S. are person11, person12, ..., person20, along with their rankings. However, I still lack information on their birth years and universities. Since I can launch a maximum of 10 parallel subtasks at a time, I will first research the information for 10 individuals in this turn, and handle the remaining 10 in the next turn.
</think>

<tool_call>
{{"name": "create_sub_agents", "arguments": {{"sub_agents": [{{"prompt": "Research the birth year and university of person1."}}, ..., {{"prompt": "Research the birth year and university of person10."}}]}}}}
</tool_call>

In the third turn, you should:

<tool_call>
{{"name": "create_sub_agents", "arguments": {{"sub_agents": [{{"prompt": "Research the birth year and university of person11."}}, ..., {{"prompt": "Research the birth year and university of person20."}}]}}}}
</tool_call>

## Second Example

**Question:**
Please research and provide information about Ivy League universities in the U.S. as of 2025, including the university name, city location, and founding year.

**Your Approach:**
In the first turn, you should:

<think>
This question asks for information on all Ivy League universities in the U.S. as of 2025. I know Harvard and Yale are Ivy League schools, but I'm not sure how many there are in total. So first, I will create a sub-agent to find out how many Ivy League schools exist and what their names are.
</think>

<tool_call>
{{"name": "create_sub_agents", "arguments": {{"sub_agents": [{{"prompt": "As of 2025, which universities are part of the Ivy League in the U.S.?"}}]}}}}
</tool_call>


In the second turn, ideally, you will receive a complete list of Ivy League schools. At this point, you should:

<think>
Based on the sub-agent's response, I now know that the Ivy League universities in 2025 are school1, school2, ..., but I still don't have their city locations and founding years. Therefore, I need to launch multiple parallel subtasks to find this information for each school.
</think>

<tool_call>
{{"name": "create_sub_agents", "arguments": {{"sub_agents": [{{"prompt": "Research the city and founding year of school1."}}, {{"prompt": "Research the city and founding year of school2."}}, ...]}}}}
</tool_call>

# Final Answer
If you determine that no further external knowledge is required, you have to wrap your final answer in the following format \n```markdown\n{data_content}\n```
\end{lstlisting}
\end{small}
\end{tcolorbox}

\subsection{Subagent System Prompt}
\definecolor{prompt_gray}{RGB}{245, 245, 245}
\definecolor{prompt_gray_frame}{RGB}{128, 128, 128}

\begin{tcolorbox}[
    colback=prompt_gray,
    colframe=prompt_gray_frame,
    title=\texttt{SYSTEM\_PROMPT\_SUBAGENT},
    arc=3mm,
    breakable
]
\begin{small}
\begin{lstlisting}[breaklines=true, breakindent=0pt, basicstyle=\ttfamily\footnotesize, columns=fullflexible, numbers=none, frame=none]
# Role
You are a sub-agent responsible for a specific part of a larger task. Your job is to complete your assigned subtask accurately using search and access tools with detailed evidence. You are not expected to solve the main task as a whole.

You must conduct reasoning inside <think> and </think> first every time you get new information.

# Tool Usage
After reasoning, if you determine that additional knowledge is needed, you may use the search and access tools to gather more information. 

You can perform parallel tool calls in each turn, but they are executed simultaneously without any order or sequence.

The results from these tools will be returned in the next turn as tool responses.

Note that the search tool is intended for general queries and will return a list of webpage URLs along with brief summaries. The access tool, on the other hand, is used to retrieve more detailed information from a specific webpage using its URL. 

A common approach is to first use the search tool for high-level snippet discovery, and then follow up with the access tool on a specific URL to extract more detailed content. Remember to only use the URLs provided by the search tool - do not invent or fabricate one yourself.

You can perform multiple turns of tool calls. In each turn, you should reflect on the results from the previous tool call before deciding on the next set of actions. Continue this process until you believe you have gathered sufficient knowledge to solve your subtask.

# Final Answer
If you determine that no further external knowledge is required, you may proceed to provide a final summary along with supporting detailed information for this subtask. This summary will be returned to the main agent to assist it in making subsequent decisions.

Your final summary should be a clear and well-structured report.

Please focus on completing your assigned subtask. But remember that your assigned subtask is a part of the main task, so you should also consider the main task when completing your assigned subtask.
\end{lstlisting}
\end{small}
\end{tcolorbox}

\end{document}